\documentclass[a4]{article}
\usepackage{isspa-spconf,amsmath,epsfig}

\begin{document}

\title{Discriminative Phoneme Sequences Extraction for Non-Native Speaker's Origin Classification}

\name{ G. Bouselmi, D. Fohr, I. Illina, J.-P. Haton\thanks{Thanks to HIWIRE project for funding.}}

\address{Speech Group, LORIA-CNRS \& INRIA, ``http://parole.loria.fr/''\\
BP 239, 54600 Vandoeuvre-l\`es-Nancy, France\\
{\small \tt \{ bousselm,fohr,illina,jph \}@loria.fr}
}


\author{}
\maketitle

\begin{abstract}

In this paper we present an automated method for the classification of the origin of non-native speakers.
The origin of non-native speakers could be identified by a human listener based on the detection of typical 
pronunciations for each nationality. 
Thus we suppose the existence of several phoneme sequences that might allow
the classification of the origin of non-native speakers.
Our new method is based on the extraction of discriminative sequences of phonemes from a
non-native English speech database. 
These sequences are used to construct a probabilistic classifier for the speakers' origin.
The existence of discriminative phone sequences in non-native speech is a significant result of this work.
The system that we have developed achieved a significant correct classification rate of 96.3\% and a significant
error reduction compared to some other tested techniques.

\end{abstract}

\section{Introduction}

The problem of non-native speaker origin classification consists in detecting the mother tong of speakers
uttering non-native speech. 
For example, the detection of the nationality of Spanish or French people uttering English words.
This is different from simple origin detection as in the latter case, the decision is taken over native speech 
(ex. English people uttering English speech vs. French people uttering French speech).
The issue we target here is closer to regional accent detection for the same native language.\\

With the recent advances in the speech recognition field, the automatic speech recognition (ASR) 
is more and more used especially in call centers. 
ASR is beneficial for both the callers and the call center companies as it allows an automated and 
fast processing with natural language and allows the reduction of the number of human operators in the repetitive
task of phone replying. Let's consider the application of a car renting call center based on automatic speech recognition.
In such case, the ASR system will interact with the customer and collect the information of his order such as
the car type, the duration of the rent, the pick point etc. 
In this case, if the origin of non-native speakers is known, an adapted ASR system can be used in order to
have better recognition accuracy.
For the latter application as for a plane ticketing call center, there is a high probability of encountering
non-native speakers.\\

The work presented here is part of the European pro- ject \textit{HIWIRE} : Human Input that Works in Real
Environments. It aims at the developments of means for helping human operators performing their duties
in real environment conditions. The \textit{HIWIRE} project consists in developing an automated system
based on ASR that assists aircraft pilots in their tasks and communications.
As communications between pilots and control operators must be in English,
the system that is under construction within \textit{HIWIRE} will inherently be confronted
with non-native English speech.
Traditional ASR systems would be inefficient in such case as their performance drops drastically
when confronted with non-native speech.
This performance drop is a well known problem (see \cite{ES000}).\\

Recent research works for non-native speech have already allowed a significant improvement in that filed.
The approaches described in \cite{ES000}, \cite{ES1} and \cite{ES4} 
allowed significative performance enhancement against non-native speech.
Nevertheless, those approaches require the knowledge of the origin of the speakers uttering the speech 
they are applied to.
Indeed, the modifications applied to the ASR system depend on both the native language and the spoken language.\\

A foreign accent classification procedure could be a great asset to any system based on speech 
recognition and confronted with non-native speakers.
Only few articles have been published concerning non-native accent classification.
For that matter, the approach developed by C. Teixeira et al. \cite{ES7} was based on HMM phone models.
It achieved 65.5\% classification rate on isolated words database of Danish, German, British,
Spanish and Italian speech.
Arslan et al. \cite{ES5} used HMM phone models and HMM word models to identify Neutral, Chinese, Turkish and
German accents. The method achieved a 68.3\% classification rate on 5 isolated words.
The approach of P. Angkitirakul et al. \cite{ES6}, based on Stochastic Trajectory Models (STM) and
Parametric Trajectory Models (PTM), performed by 40.6\% classification rate in supervised mode.\\

In the next section, we will describe the extraction of discriminative phone sequences for each foreign language and will
sketch the decision process based on conditional probabilities.
In section \ref{Experiments}, we will describe the tests that we have carried and discuss their results.
We will also discuss the future research work.
Finally, we will end with a brief conclusion.

\section{Foreign accent detection}

It is well known that non-native speakers might produce pronunciation errors when uttering foreign speech (\cite{ES000}, \cite{ES1}, \cite{ES4}, \cite{ES10}, \cite{ES11}).
These errors are due to the phonological and articulatory properties of both the spoken and native languages.

For instance, some phonemes of the spoken language (SL) might not exist in the native language (NL) of the speaker.
The speaker may replace these phonemes by some acoustically close phonemes of his NL.
For instance, in French, diphthongs do not exist and some French speakers pronounce
instead a sequence of two French phones.

These non-native pronunciation error depends on the pair of spoken and native languages.
P. Ladefoged et al. \cite{ES10} and R.J. Jeffers et al. \cite{ES11} have depicted in their work a list of common phone uttering errors
made by various groups of non-native speakers for the English language (French, Italian, Greek etc.).
Indeed, speakers from the same origin are very likely to commit the same pronunciation errors as they share the
same native language, and thus the same articulatory and phonological mechanisms.
Human listeners rely on those common errors as hints and tips to decide on the origin of non-native speakers.
In the work of Arslan et al. \cite{ES5} and Angkitirakul et al. \cite{ES7}, human listener achieved 54\% and
84\% in foreign speech classification.\\

Our approach described here is based on that feature.
We suppose the existence of discriminative uttering structures at the phonetic level that
are shared among speakers from a particular origin when they speak a foreign language.
In other words, we suppose that speakers from a particular origin $X$ utter some discriminative
sequences of phonemes when they speak a foreign language $Y$.\\

We suppose that for a set of origins $L = \{L_1 .. L_n\}$ and a foreign language $F$, there exist
sets of phoneme sequences $S_1 .. S_n$ corresponding to the origins $L_1 .. L_n$ (respectively)
that might discriminate the native languages of $L_1 .. L_n$ speakers when they utter $F$ speech
($S_i = \{ s_{i,1} .. s_{i,{k_i}} \}$, and $s_{i,m}$ are sequences of phones).

\subsection{Discriminative phoneme sequences extraction}
\label{discriminative_extraction}

In order to better model the non-native speech, we have chosen to use the phone acoustic models (HMM) of
all the native languages $L_1 .. L_n$ (the models are noted $M_1..M_n$ respectively).
Let the non-native database $B = \bigcup_{i=1}^n B_i$, where $B_i$ is the part of the database composed of $L_i$ speakers
uttering $F$ speech.
First, the native phone models are adapted on the respective non-native database:
i.e. the models $M_1$ are adapted on the $B_1$ resulting in $M'_1$, and so on.
Then, to extract these discriminative phone sequences, we perform a phonetic recognition with a phonetic pool
$M = \bigcup_{m=1}^n M'_m$ on each of the non-native databases.
For each native language $L_i$, we count the occurrences of all the phone sequences having a maximum length of $max_{p}$ phones
in the phonetic recognition results.
The number of occurrences of the phone sequences is normalized against the number of sentences that compose each non-native parts of the database ($B_1..B_n$).\\

This processing results in sets of preliminary phone sequences with their normalized number of occurrences for each language $L_i$.
Those sets are noted $S'_i = \{ s_{i,1} .. s_{i,{k'_i}} \}$, and the number of occurrences is noted $n_i(s)$ for a sequence s (for a language $L_i$).
The next step consists in retaining the sets of discriminative sequences $S_i$.
For $L_i$, a sequence $s \in S'_i$ is considered discriminative only if it verifies the equation \ref{equa_discrim}.

\begin{equation}
\label{equa_discrim}
n_i(s) >= \alpha * n_k(s), \forall k \ne i
\end{equation}
where $\alpha$ is a discriminant factor, $\alpha\ge 1$.\\

Knowing the $S_i$ sets and the counts of appearances of each of their sequences, some probabilities can be computed.
All the following probabilities are conditional probabilities conditioned by the acoustic models $M$ and the sets $S_1..S_n$.
For readability reasons, we will omit these conditions in the probabilities notations.
The maximum likelihood (ML) probability $P(L_i)$, $P(s)$ and $P(s/L_i)$ are 
computed as follows :

\begin{equation}
\label{equa_p_L}
P(L_i) = \frac{  \Sigma_{m=1}^{k_i} n_i(s_{i,m})}
{ \Sigma_{l=1}^n \Sigma_{m=1}^{k_l} n_l(s_{l,m}) }
= \frac{\Sigma_{x \in S_i} n_i(x) }{ \Sigma_{l=1}^n \Sigma_{x \in S_l} n_l(x) }
\end{equation}

\begin{equation}
\label{equa_p_s}
P(s) = \frac{  \Sigma_{l=1}^{n} n_i(s) }
{ \Sigma_{l=1}^n \Sigma_{m=1}^{k_l} n_l(s_{l,m}) }
= \frac{\Sigma_{l=1}^{n} n_i(s)}
{ \Sigma_{l=1}^n \Sigma_{x \in S_l} n_l(x) }
\end{equation}

\begin{equation}
\label{equa_p_s_l_i}
P(s / L_i) = \frac{n_i(s)}{ \Sigma_{m=1}^{k_i} n_i(s_{i,m}) }
= \frac{n_i(s)}{ \Sigma_{x\in S_i} n_i(x) }
\end{equation}

Using the bayes rule and the equations \ref{equa_p_L}, \ref{equa_p_s} and \ref{equa_p_s_l_i}, the conditional probability of a language $L_i$ knowing a sequence $s$ can be computed as in the equation \ref{equa_p_L_s}.
\begin{equation}
\label{equa_p_L_s}
P(L_i / s) = \frac{P(s/L_i) * P(L_i) }{P(s)} $$ \\ $$
=\frac{
\frac{n_i(s)}{ \Sigma_{x\in S_i} n_i(x) } *
\frac{\Sigma_{x \in S_i} n_i(x) }{ \Sigma_{l=1}^n \Sigma_{x \in S_l} n_l(x) }
}
{
\frac{\Sigma_{l=1}^{n} n_i(s)}{ \Sigma_{l=1}^n \Sigma_{x \in S_l} n_l(x) }
}
=
\frac{n_i(s)}{ \Sigma_{l=1}^{n} n_l(s) }
\end{equation}

The conditional probability of a language $L_i$ knowing a list of sequences $O = \{ s_1 .. s_h \}$ can be computed as in equation \ref{equa_p_L_O} using the bayes rule, the equations above and the hypothesis that sequences of $O$ are independent.
The hypothesis of independence of the sequences is not true.
Nevertheless, this hypothesis must be assumed in order to compute this probability.
Indeed, determining the interrelations between sequences of phones might prove to be impossible to compute with regards to the small size of our database.

\begin{equation}
\label{equa_p_L_O}
P(L_i / O) = \frac{P(O/L_i)P(L_i) }{P(O)} = \frac{P(s_1..s_h/L_i)P(L_i) }{P(s_1..s_h)} $$ \\ $$
=\frac{P(L_i)\prod_{m=1}^h P(s_m/L_i)}{\prod_{m=1}^h P(s_m)} $$ \\ $$
=P(L_i)^{1-h}\prod_{m=1}^h \frac{ P(s_m/L_i)P(L_i)}{P(s_m)} $$ \\ $$
=P(L_i)^{1-h}\prod_{m=1}^h P(L_i/s_m)
\end{equation}

\subsection{Classification of a speaker}
\label{Classification_of_a_speaker}

In order to detect the origin of a speaker $X$, some of its recorded utterances must be analyzed.
First, a phonetic recognition is performed on those sentences using a the models $M$ described in \ref{discriminative_extraction}.
All the sequence of phones that appear in the sets $S_1 .. S_n$ are retained in a list $O = \{s_1..s_h\}$.
The speaker $X$ is classified in the $L_i$ native language as in equation \ref{equa_classification}.
\begin{equation}
\label{equa_classification}
L_i = argmax_{l=1..n} \{ P(L_l / O) \}
\end{equation}

Another local decision approach can be adopted.
Instead of collecting all the sequences of phones from the phonetic recognition (see last paragraph) that appear in $S_1 .. S_n$ in a single list $O$,
separate lists $O_1 .. O_n$ corresponding to the sequences that appear in $S_1 .. S_n$ (respectively) could be made up.
I.e., the list $O_l$ corresponds to all the sequences of phones observed in the phonetic recognition and that appear in the set $S_l\ (l=1..n)$.
The decision is then made over the probabilities of the languages knowing the lists $O_1.. O_n$, i.e. $P(L_l/O_l), \forall l=1..n$.
In this classification approach, the probabilities of each language must be normalized over the number of sequences of each list $O_1 .. O_n$ in order to allow the comparison
between them.
Besides, any language decider that has a too small corresponding list of observations should be ignored, 
i.e., for a list $O_i$, if exists $k$ verifying $card(O_k) \ge \beta\ card(O_i)$, the classifier $L_i$ is ignored ($\beta$ is a factor).
If we note $I$ the set of language indices that are not ignored, the speaker $X$ is classified in the language $L_i$ if the equation \ref{equa_classification_2} is verified.

\begin{equation}
\label{equa_classification_2}
L_i = argmax_{l \in I} \{ P(L_l / O_l)^{\frac{1}{card(O_l)}} \}
\end{equation}

\section{Experiments}
\label{Experiments}
\subsection{Experimental conditions}

Our tests have been carried out on the \textit{HIWIRE} non-native speech database.
This database have been tested in the approaches presented in \cite{ES000} and \cite{ES00}.
It is composed of 81 speakers: 31 French, 20 Greek, 20 Italian and 10 Spanish speakers.
Each of those speakers reads 100 English sentences.
The used grammar is a strict command language composed of 134 words.
This grammar is used by aircraft pilots when communicating with airport control agents.
The speech was recorded in 16 bits and 16 kHz format.
We chose an MFCC parametrization with 13 coefficients and their first and second time derivatives.
The acoustic models are 3 states HMMs (Hidden Markov Models) with 128 Gaussian mixtures and diagonal covariance
matrices.

\subsection{Tests and results}

In ours tests, we have used 39 French, 33 Greek, 32 Spanish and 49 Italian monophone HMM models trained
on native speech databases (respectively).
As described in section \ref{discriminative_extraction}, those models were adapted on \textit{HIWIRE} non-native
database.
I.e., the French models were adapted on all the French speakers, etc.\\

The extraction of the phone sequences was done following the ``leave one out'' scheme.
For instance, when testing a French speaker $X$, the discriminative phone sequences of the French language
are extracted using all the French speakers except $X$. And in that example, the significant sequences
of the other languages are extracted using all the respective speakers.\\

In our preliminary tests, we have chose some threshold values as follows :
\begin{itemize}
\item[-] $\alpha=4$ : the significance factor (see section \ref{discriminative_extraction}).
\item[-] $max_p=3$ : the maximum length of a phone sequence (number of phones in the sequence).
\item[-] 50 as the minimum occurrences count per speaker for a sequence to be eligible as discriminative.
\item[-] 30 as the maximum discriminative sequences count per language.
\item[-] $\beta = 2.5$ : see section \ref{Classification_of_a_speaker}.
\end{itemize}

Table \ref{table_results} shows the confusion matrix in terms of speakers percentage for the global decision
matrix. The accuracy achieved is 96.29\% of correct speaker classification.
Table \ref{table_results_local_decision} shows the confusion matrix in terms of speakers percentage for the local decision
matrix. The accuracy achieved is 87.65\% of correct speaker classification.

\begin{table}
\caption{\label{table_results}
{\it Confusion matrix for the global decision method. The classification rate is 96.29\%.}}
\begin{tabular}{|c|c|c|c|c|}
\hline
	& French & Greek & Italian & Spanish \\ \hline
French  & 100.0  & 0.0   & 0.0     & 0.0     \\ \hline
Greek	& 5.0    & 95.0  & 0.0     & 0.0     \\ \hline
Italian	& 0.0    & 5.0   & 95.0    & 0.0     \\ \hline
Spanish & 0.0    & 0.0   & 10.0    & 90.0    \\ \hline
\end{tabular}
\end{table}

\begin{table}
\caption{\label{table_results_local_decision}
{\it Confusion matrix for the local decision method. The classification rate is 87.65\%.}}
\begin{tabular}{|c|c|c|c|c|}
\hline
	& French & Greek & Italian & Spanish \\ \hline
French  & 77.1   & 0.0   & 0.0     & 12.9    \\ \hline
Greek	& 5.0    & 85.0  & 5.0     & 5.0     \\ \hline
Italian	& 5.0    & 5.0   & 85.0    & 5.0     \\ \hline
Spanish & 0.0    & 0.0   & 0.0     & 100.0   \\ \hline
\end{tabular}
\end{table}

\subsection{Discussion and future work}

The tests presented here are only preliminary and we intend to further investigate and tune the parameters of this method.
The effect of some parameters like the significance factor $\alpha$, the maximum number of sequences per language and the minimum appearance count for sequen-
ces will be investigated in our future work.
Besides, we will test the use of English acoustic models adapted on non-native speech instead of native models.
This might avoid the inconvenience of collecting all the acoustic models of all non-native language that will be classified.\\

The potential of native language classification based
on discriminative phone sequence might be great. Indeed,
we have tested other mother tong detection (over nonnative
speech) techniques inspired from the state of the
art. We have tested a global GMM classification where a
GMM was trained for each native language. We have also
tested three HMM based approaches using TIMIT context
independent phonemes. We have adapted the TIMIT
phone models on the French, Greek, Spanish and Italian
databases in a supervised fashion.
The best result obtained with those methods
is only 84\% on the same HIWIRE database we have used. The
global decision approach described above achieved a significantly
better result of 96.3\%, giving an error reduction
of 76.9\% (relative).\\

Another significant result of our work is the existence of discriminative phone sequences -or syllabic realizations-
in non-native speech. Those phone sequences can be relied on to classify the origin of non-native speakers.

\section{Conclusion}
\label{Conclusion}

In this paper, we presented a novel approach for the detection of the mother tong of non-native speakers based on discriminative phone sequences.
We have determined that there exists some discriminative phone sequences in non-native speech that could help in the mother tong detection.
The preliminary results we obtained show a great potential for this technique: 96.3\% correct classification rate.
Our method will be further tested and tuned for the best classification results.

\section{Acknowledgments}

This work was partially funded by the European project \textit{HIWIRE}
(\emph{Human Input that Works In Real Environments}), contract number 507943,
\emph{sixth framework program, information society technologies}.


\end{document}